\documentclass{article}
\usepackage{fourier}
\usepackage[margin=1in]{geometry}

\usepackage[utf8]{inputenc}
\usepackage{amsmath,amssymb}
\usepackage[round]{natbib}
\usepackage{microtype}
\usepackage{graphicx}
\usepackage{subfigure}
\usepackage{booktabs} 

\usepackage{todonotes}

\usepackage{enumitem}

\begin{document}

\title{Depth and nonlinearity induce implicit exploration for RL}
\author{
Justas Dauparas$^{1,2}$\footnote{Part of this work was done while Justas was a Research Intern at Microsoft Research Cambridge.},\quad
Ryota Tomioka$^{2}$,\quad
Katja Hofmann$^{2}$\vspace{5mm}\\
${}^1$University of Cambridge \& 
${}^2$Microsoft Research, Cambridge, UK
}
\maketitle
\begin{abstract}
The question of how to explore, i.e., take actions with uncertain outcomes to learn about possible future rewards, is a key question in reinforcement learning (RL). Here, we show a surprising result: We show that Q-learning with nonlinear Q-function and no explicit exploration (i.e., a purely greedy policy) can learn several standard benchmark tasks, including mountain car, 
equally well as, or better than, the most commonly-used $\epsilon$-greedy exploration. We carefully examine this result and show that both the depth of the Q-network and the type of nonlinearity are important to induce such deterministic exploration.
\end{abstract}

\section{Introduction}

Reinforcement learning (RL) is a systematic approach to learning in sequential decision problems, where a learners' future task performance depends on its past actions. In such settings, learners have to explore, meaning they have to take actions with uncertain outcomes, to facilitate learning about the consequences of such actions.

The question of how to best explore is a key open question in RL. Here, we specifically address this question from an empirical perspective, and investigate how to explore in a way that leads to sample efficient learning in deep RL, i.e., reinforcement learning with value function approximators that are parameterized as powerful neural networks.

We present a surprising finding: in this setting, good approximate value functions can be learned without any explicit exploration. In fact, we find that in several cases learning without explicit exploration is equally or more sample efficient than the most-commonly used $\epsilon$-greedy exploration scheme on several standard benchmark tasks.
%
%
We present additional results that suggest a likely role of model structure (network depth and nonlinearity) in inducing such implicit exploration. We believe that our insights have strong practical implications and open up a novel line of research towards understanding exploration in deep RL.

\section{Methods}

We briefly outline the components to our approach that form the basis of our investigation.
We assume a standard formulation of RL as learning in a Markov Decision Process, where the learner is tasked to find an optimal policy $\pi^*$. For any given policy $\pi$ the Q-value, also called state-action value, can be written as $Q^{\pi}(a,s) := \mathbb{E}[ r(s, a) + \sum_{t = 1}^{\infty} \gamma^t r(s_t, a_t)]$, i.e., the expected discounted (with discount factor $\gamma$) cumulative reward from taking action $a$ in state $s$ and following policy $\pi$ thereafter. An optimal policy achieves optimal Q-values $Q^* := max_{\pi} Q(s, a)$. 

\paragraph{Learning approach (DDQN)}
Q-learning-based approaches estimate $Q^*$ using an iterative approach that bootstraps estimates of $Q(s,a)$ from those of subsequent states $s^\prime$, using the recursion $Q(s,a) = r(s,a) + \gamma max_{a^\prime} Q(s^\prime, a^\prime)$. In approaches based on deep Q-learning \citep{Mnih2015}, Q-value estimates are parameterized by a deep neural network, and trained using stochastic gradient descent using interaction data obtained through interaction with an environment using a behavior policy. In Double DQN (DDQN \citep{vanhasselt2016deep}), gradient updates minimize the squared loss $\|Q(s,a; \theta_t) - r(s,a) - Q(s^\prime, argmax_{a^\prime} Q(s^\prime, a^\prime; \theta_t); \theta_t^\prime)\|^2$, where the parameters of the Q function are denoted by $\theta$ and we explicitly distinguish between model parameters $\theta$ and target parameters $\theta^\prime$. Stochastic updates are computed on mini-batches sampled from a replay buffer, a record of past experience.

\paragraph{Exploration}
We contrast a greedy behavior policy with the standard $\epsilon$-greedy approach \citep{sutton1998reinforcement}. A greedy policy selects actions $a_{\theta}^* = argmax_a Q(a,s; \theta)$. In $\epsilon$-greedy, actions are sampled uniformly at random with exploration rate $\epsilon$, while the greedy action is selected with probability $1-\epsilon$. Following common practice \citep{Mnih2015}, we decay the exploration rate over time.


\begin{figure*}[t]
\subfigure[$\epsilon=0$]{
\includegraphics[width=.33\textwidth]{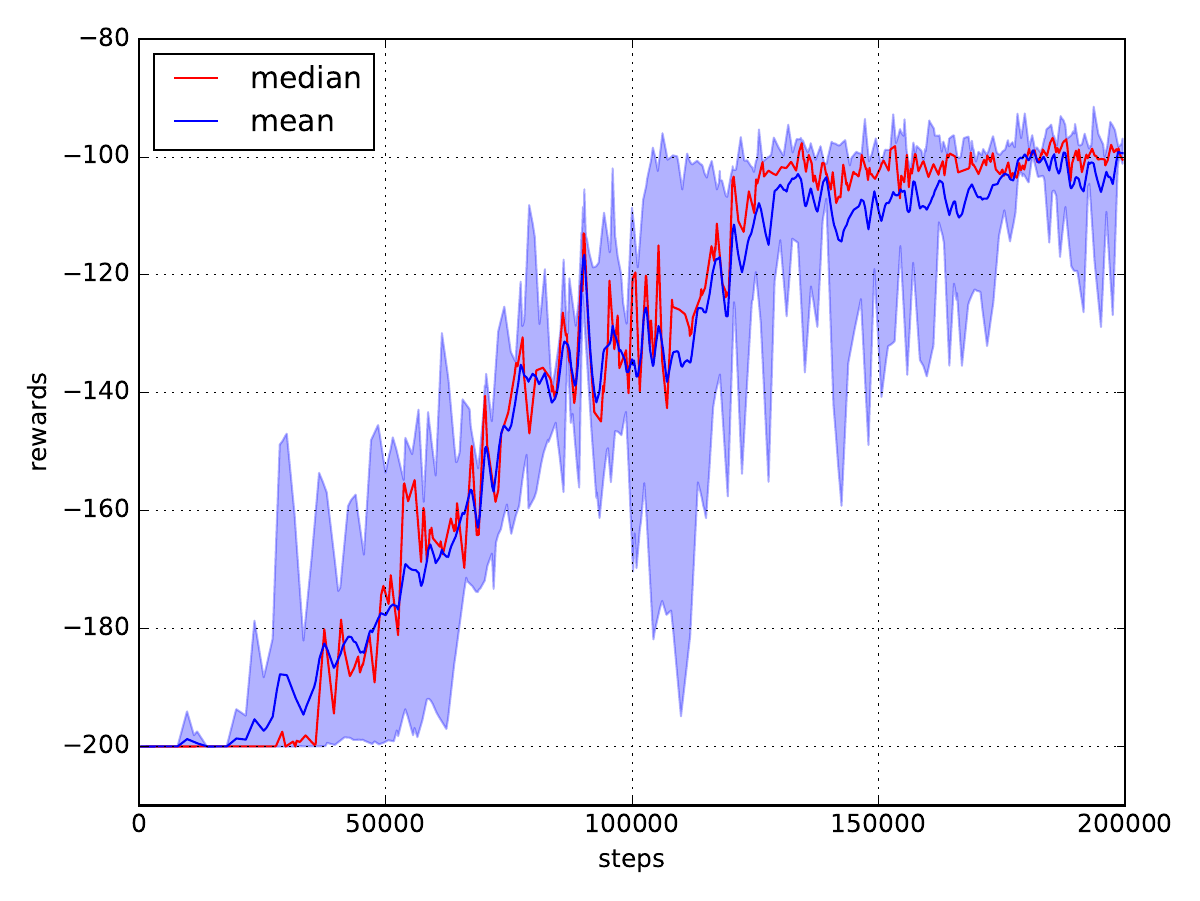}}
\subfigure[$\epsilon$ decayed in 25k steps]{
\includegraphics[width=.33\textwidth]{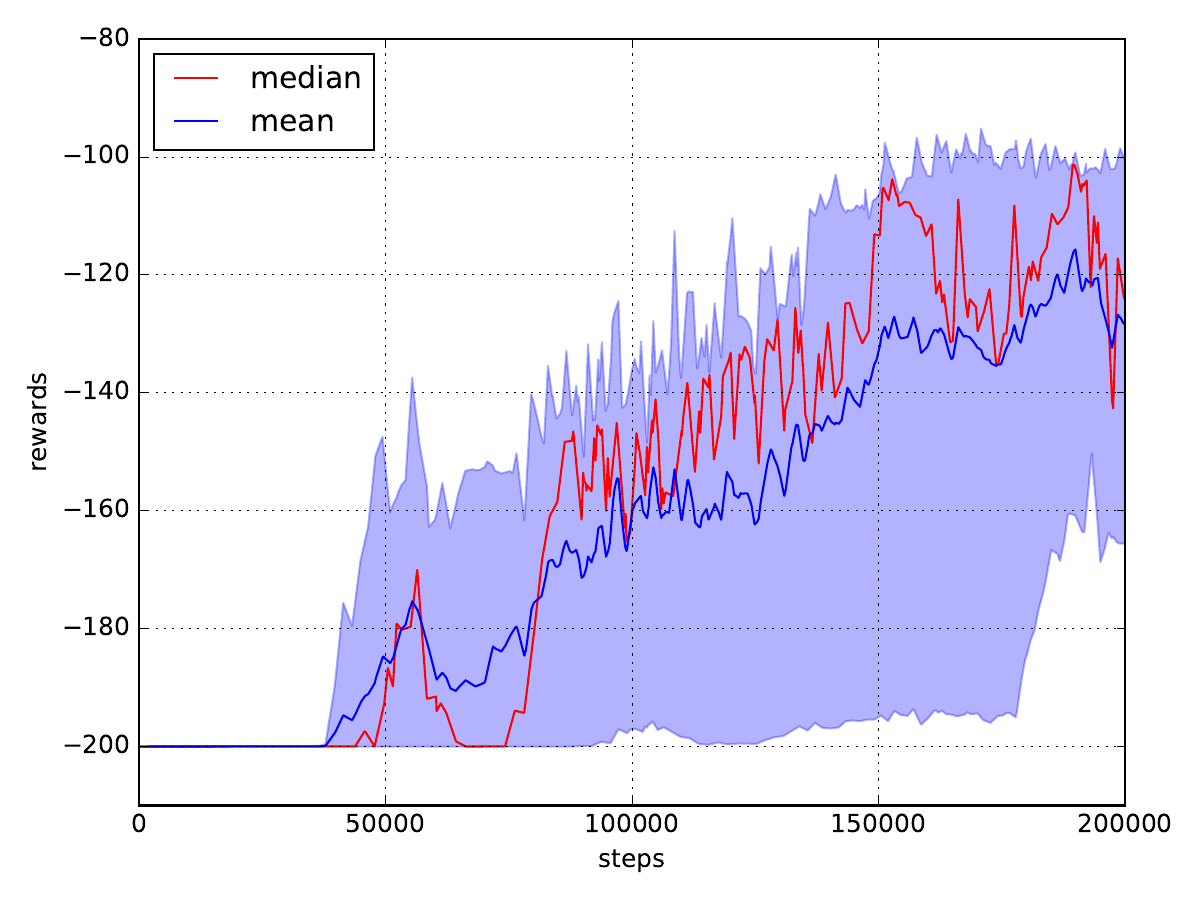}}
\subfigure[$\epsilon$ decayed in 100k steps]{
\includegraphics[width=.33\textwidth]{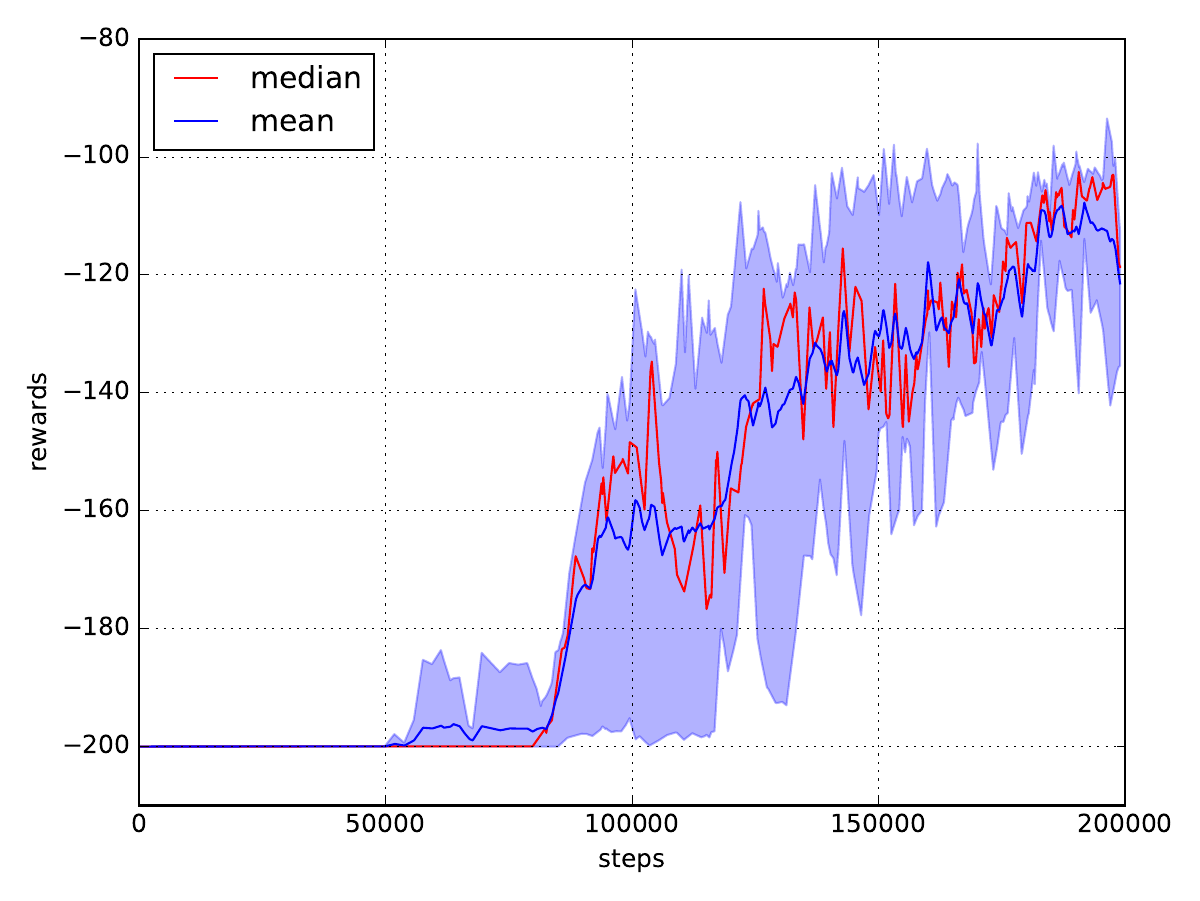}}
\caption{Comparison of no explicit exploration ($\epsilon=0$) to linearly decaying $\epsilon$ on the mountaincar-V0 task (5 random seeds).}
\label{fig:decay}
\end{figure*}

\begin{figure*}[t]
\subfigure[cartpole-v0, $\epsilon=0$]{
\includegraphics[width=.25\textwidth]{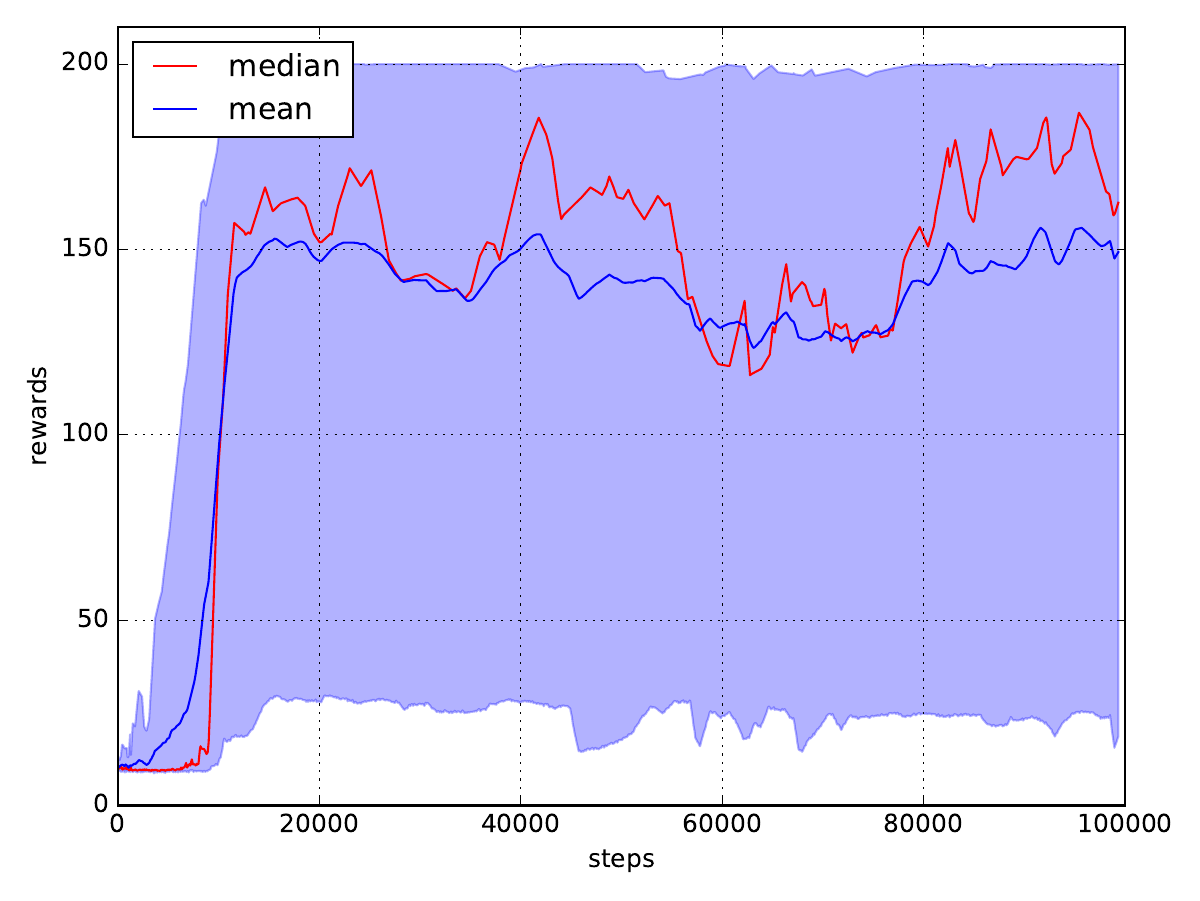}
}~\subfigure[cartpole-v0, $\epsilon$ dec. in 10k steps]{
\includegraphics[width=.25\textwidth]{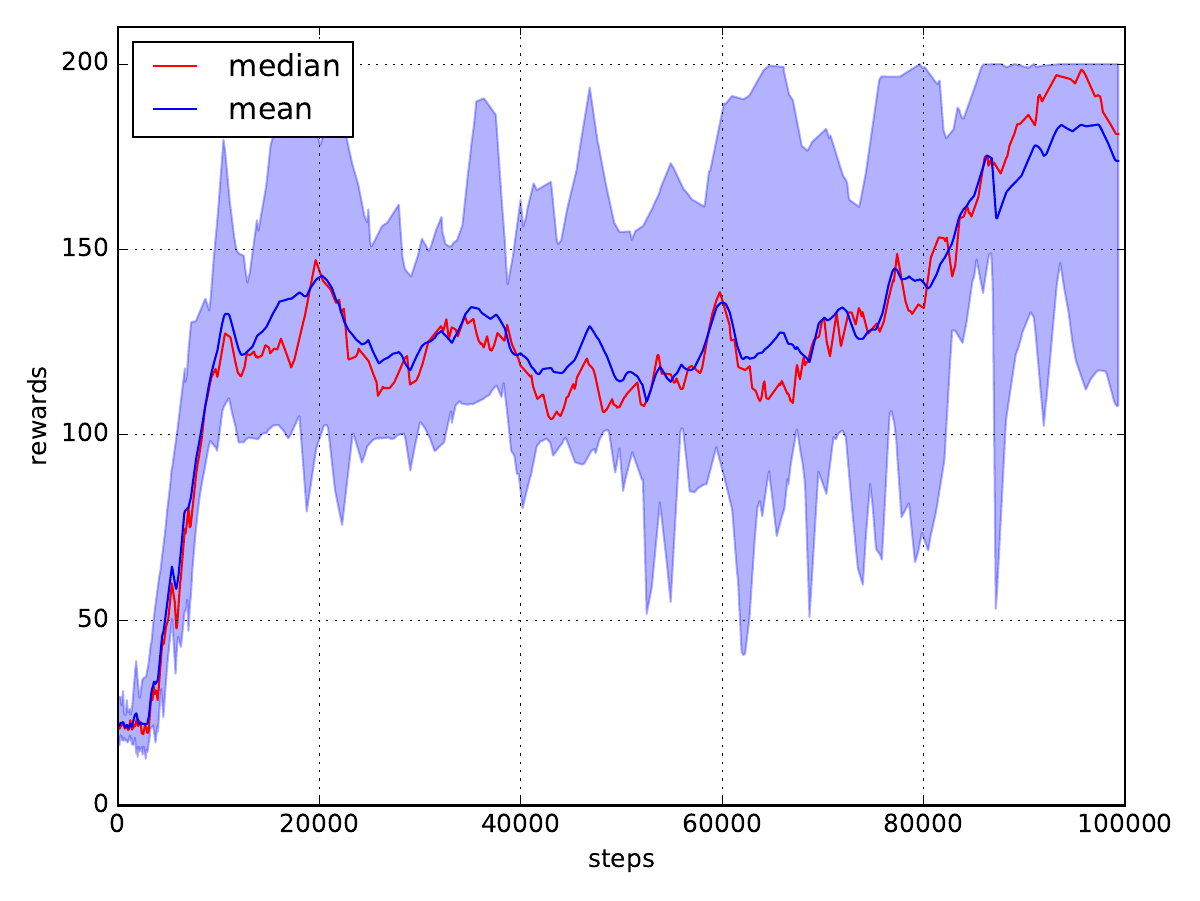}
}~\subfigure[acrobat-v1, $\epsilon=0$]{
\includegraphics[width=.25\textwidth]{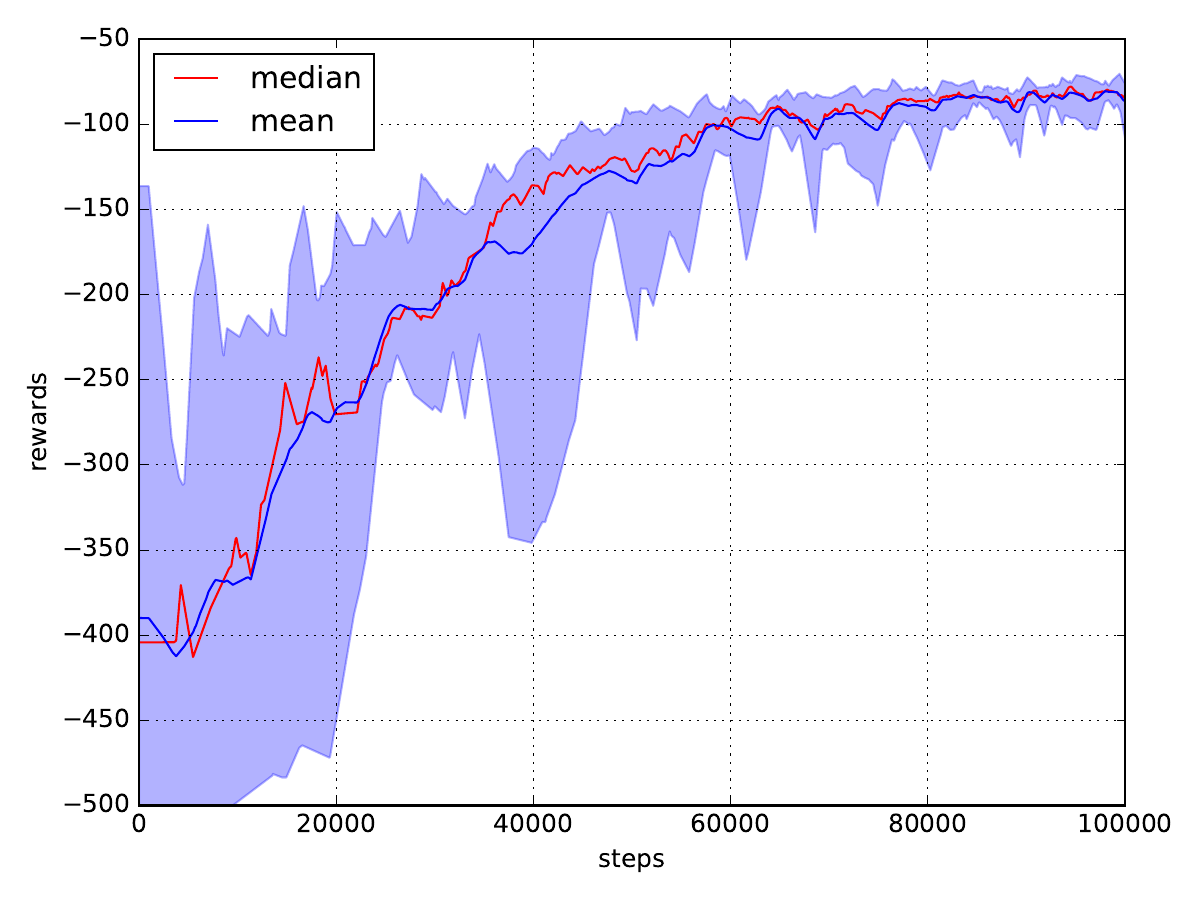}
}~\subfigure[acrobat-v1, $\epsilon$ dec. in 10k steps]{
\includegraphics[width=.25\textwidth]{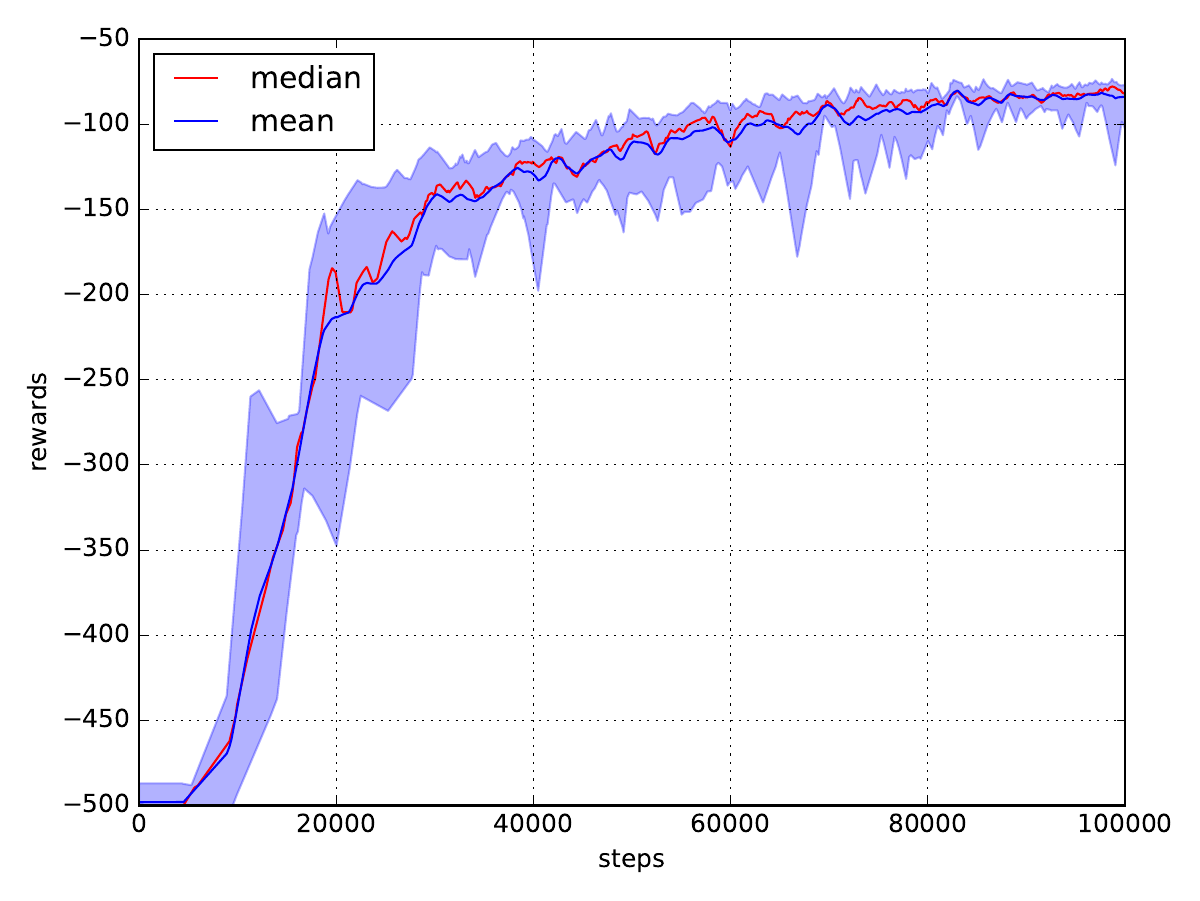}}
\caption{cartpole-v0 and acrobat-v1 tasks (10 random seeds).}
\label{fig:cartpole-acrobat}
\end{figure*}


\paragraph{Tasks}
We use the following OpenAI-gym \citep{brockman2016openai} tasks: mountaincar-v0, cartpole-v0, and acrobat-v1. These are common RL benchmarks \citep{duan2016benchmarking}.

\paragraph{Hyper-parameters}
For the experiments on mountaincar-v0, we used a replay buffer of size 200k, batch size 256, discount factor $\gamma=0.99$; we used a Q-network with two ReLU hidden layers of 128 units each, unless stated otherwise; for optimization, we used Adam \citep{kingma2014adam} with $\alpha=5\cdot 10^{-4}$; the target network was updated every 1000 steps. For cartpole-v0 and acrobat-v1 we used replay buffer size 50k; all other parameters were the same.


\section{Results}

In Figure \ref{fig:decay}, we plot the reward statistics  (mean, median, 2\%- and 98\%-pecentiles) obtained by running DDQN on the mountaincar-v0 task with (a) no explicit exploration ($\epsilon=0$) (b) linear decay of the exploration rate $\epsilon$ from 1 to 0 in 25k steps and (c) linear decay in 100k steps. 5 independent random seeds were used to obtain the statistics.
The plots show that the agent without explicit exploration ($\epsilon=0$) can solve the task equally well, or even slightly better than, standard exploration strategies. We confirmed similar results on cartpole-v0 and acrobat-v1 (Fig.~\ref{fig:cartpole-acrobat}). 

How can an agent explore without randomness? Note that all the above environments are deterministic except for the initial states. If it is not the environment or the stochasticity in the behavior policy, {\em it must be some property of the Q-network that is inducing the exploration}.

\begin{figure*}[tb]
\begin{center}
\includegraphics[width=\textwidth]{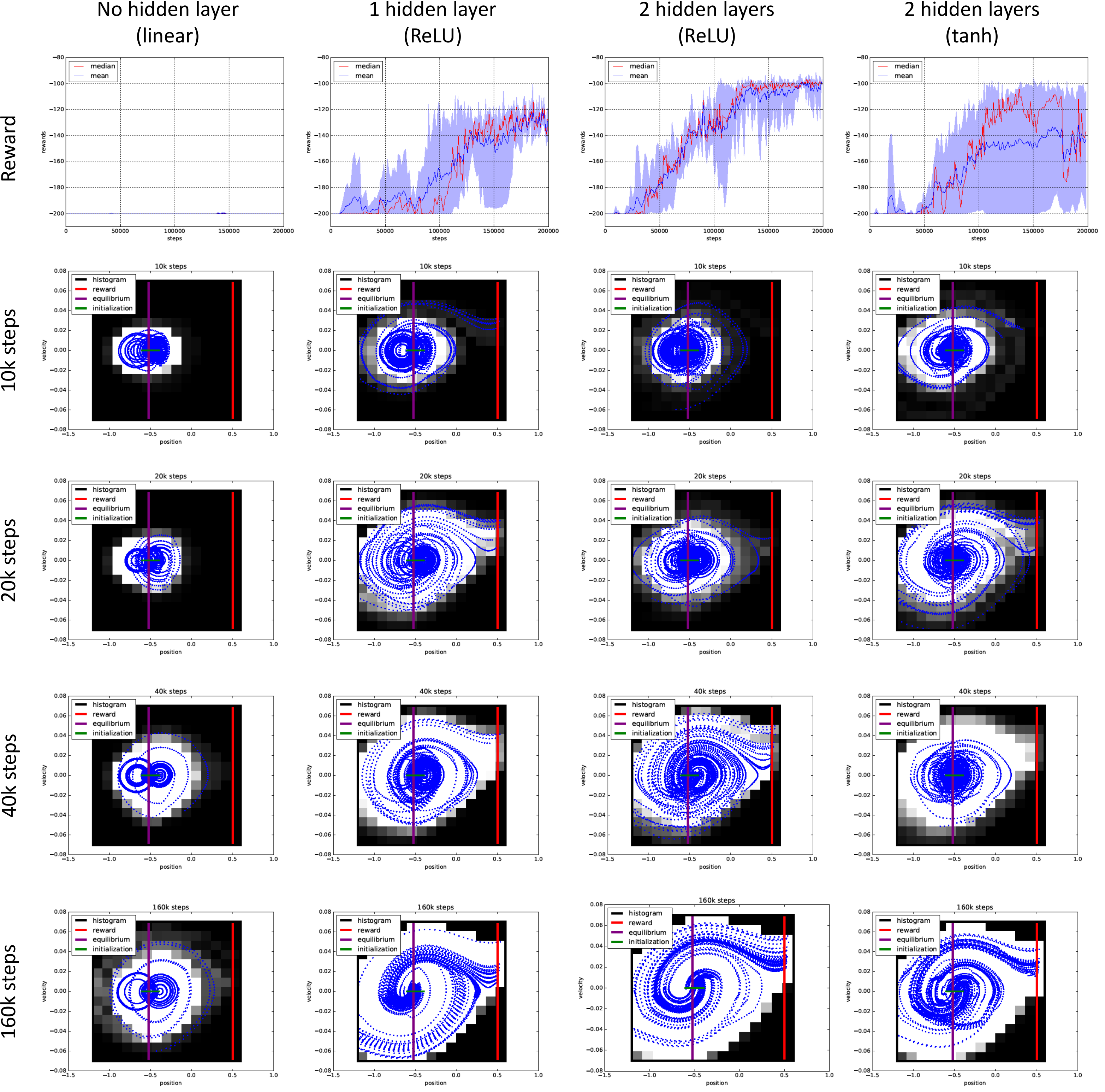}
\caption{Reward (5 random seeds) and trajectories for different Q-network architectures.}
\label{fig:arch}
\end{center}
\end{figure*}

To understand what is inducing the exploration for the mountaincar-v0 task, we carried out further experiments with the following Q-network architectures (see Fig.~\ref{fig:arch}):

\begin{enumerate}
\item Linear (no hidden layer);
\item 1 hidden layer with 128 ReLU units;
\item 2 hidden layers with 128 ReLU units in each layer (original setting);
\item 2 hidden layers with 128 tanh units in each layer.
\end{enumerate}
All results were obtained with $\epsilon=0$. Within each column, we plot the reward statistics (as above), and phase space diagrams showing the 1000 state transitions leading up to 10k steps, 20k steps, 40k steps, and 160k steps. The trajectories are superimposed on top of the histograms of the state visit frequencies colored from black (zero) to white (more than 100). The red vertical lines indicate the goal states.

In the first column of Fig.~\ref{fig:arch}, 
we can see that without any nonlinearity, the agent was not able to reach the goal state  even once and consequently did not learn the task at all, although we believe that a linear agent is sufficient to solve this task \citep{mania2018simple}.

By contrast, we can see in the second column that the agent is able to solve the task with a single ReLU hidden layer of size 128.
We have also experimented with two fully-connected layers without nonlinearity or just one fully-connected layer initialized with large weight initialization scale, but none of them were as successful as the networks with ReLU nonlinearities. The original setup of two hidden layers (third column) seem to be slightly better than one hidden layer. The last column shows the same result for two hidden layers with the {\em tanh} nonlinearity. The reward curve appears slightly noisier than for the ReLU activations, but this may be due to the high variance.

\begin{figure}[tb]
\subfigure[Random {\em nonlinear} Q-function as a controller.]{
\includegraphics[width=.5\textwidth]{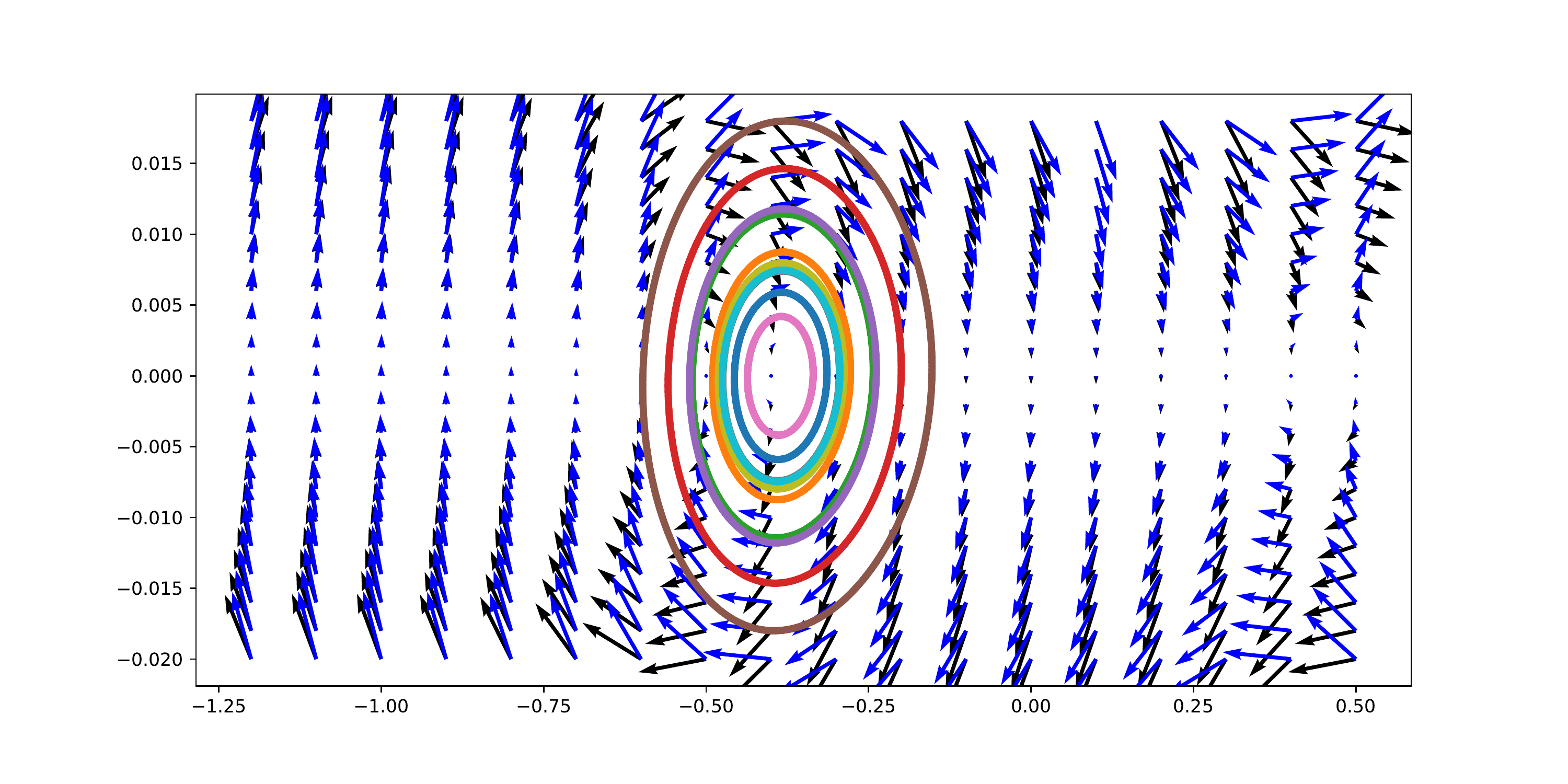}
\label{fig:init-3layers}
}~\subfigure[Random {\em linear} Q-function as a controller.]{
\includegraphics[width=.5\textwidth]{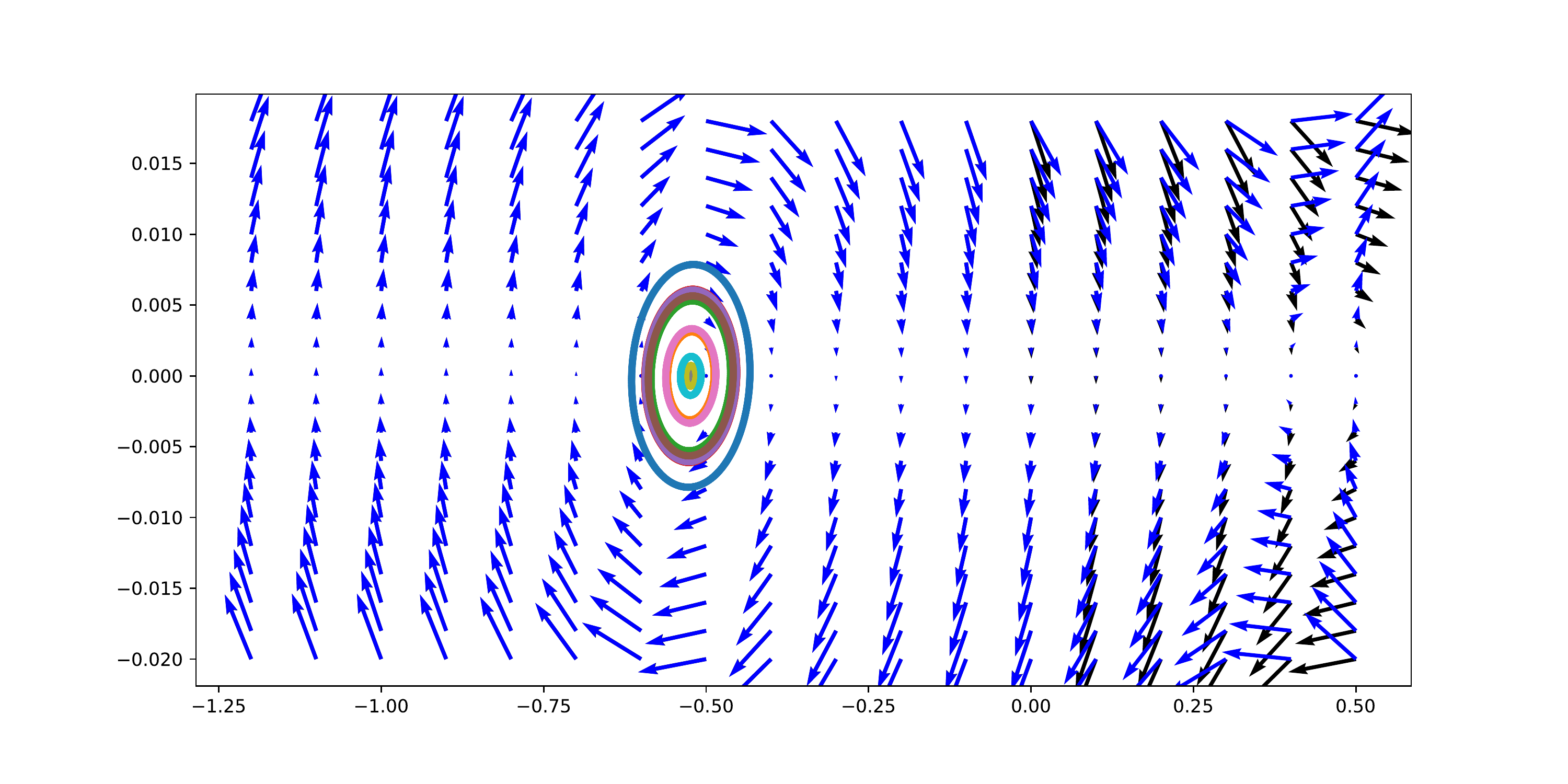}
\label{fig:init-1layer}
}
\caption{Vector fields in the phase space of the mountaincar-v0 task with and without the random Q-function as a controller. Blue: with the controller. Black: uncontrolled system.}
\end{figure}

\section{Discussion and Conclusion}
\paragraph{Can deterministic exploration be an alternative to random exploration?}
Deterministic exploration is attractive because it would avoid the unnatural dithering behavior often observed with $\epsilon$-greedy and other stochastic exploration strategies.
From a control theory perspective, an easy way to induce exploration is to destabilize the underlying system. For example, a small inverse damper term (i.e., an acceleration proportional to the speed) would be sufficient for the mountain car task, because success does not depend on the speed at which the goal state is reached. However, this is not the case for other benchmark tasks (e.g., acrobat-v1) and it would be a bad idea for real-world systems. Another way to induce deterministic exploration would be to induce chaotic dynamics. For example, for acrobat-v1, it is enough for the controller to compensate for  gravity. However, in both cases it is unlikely that a randomly drawn initial Q-function behaves like an inverse damper term or a gravity compensator.
In this paper we did not design an optimal deterministic exploration behavior but we demonstrated that such an behavior can be induced by the network architecture.

\paragraph{What is the role of the nonlinearity?}
We plot the vector fields of the moutaincar-v0 task with and without a randomly initialized Q-function with two hidden layers as a controller in Fig.~\ref{fig:init-3layers}. All weight matrices were initialized using Glorot initialization \citep{glorot2010understanding} and all bias terms were initialized to zero. We also plot 10 trajectories from random initial states. The same plot with a linear Q-function is shown in Fig.~\ref{fig:init-1layer}. Comparing the two plots we notice that the nonlinear Q-function can modify dynamics in multiple regions of the phase space (e.g., area around position$=-0.5$ and $0.5$), whereas the linear Q-function tends to focus on one region. However, as we can see in the trajectories in Fig.~\ref{fig:init-3layers}, random Q-network initialization itself is not enough to induce dynamics that can reach the goal state. Therefore we hypothesize that it is the combination of optimistic initialization \citep{sutton1998reinforcement} and the flexibility of the Q-function that is inducing exploration. That is, when the area around the initial state is explored and observed to be fruitless, a flexible nonlinear agent can still maintain optimism in the states and actions that it has not seen, whereas a linear agent may incorrectly extrapolate that there is no reward in those states.

\paragraph{Limitations} We note several limitations. First, stochasticity may be induced by initial states, as they are randomly sampled from $[-0.6, -0.4]$ in the moutaincar-v0 task. However, since the goal state cannot be reached from any of the states in this interval, this is unlikely to play a major role. 

Second, we have not fully studied how the optimistic initialization interacts with the nonlinearity of the Q-function. For mountaincar-v0, it would be interesting to experiment with an inverse reward that is positive at the goal and zero otherwise. However, optimistic initialization applies to acrobat-v1 and moutaincar-v0 (all rewards are negative) but not to cartpole-v0 (positive rewards), a task where DDQN with $\epsilon=0$ can learn as well as the $\epsilon$-greedy approach. 

\section{Discussion}

In this note we have shown that competitive performance on standard RL benchmarks can be achieved without explicit exploration when deep neural networks are used as function approximators in Q-learning. Our analysis suggests that both network depth and nonlinearity play a role by inducing optimism without overgeneralization.
While we have mainly focused on the aspect of optimism induced by a deterministic policy, another important aspect is understanding the role of uncertainty. We believe that combining uncertainty quantification \citep[e.g., bootstrapped DQN][]{osband2016deep} with deterministic exploration could be an interesting alternative to  standard stochastic exploration.

\clearpage
\bibliographystyle{named}
\bibliography{exploration}

\begin{thebibliography}{}

\bibitem[\protect\citeauthoryear{Brockman \bgroup \em et al.\egroup
  }{2016}]{brockman2016openai}
Greg Brockman, Vicki Cheung, Ludwig Pettersson, Jonas Schneider, John Schulman,
  Jie Tang, and Wojciech Zaremba.
\newblock Openai gym.
\newblock {\em arXiv:1606.01540}, 2016.

\bibitem[\protect\citeauthoryear{Duan \bgroup \em et al.\egroup
  }{2016}]{duan2016benchmarking}
Yan Duan, Xi~Chen, Rein Houthooft, John Schulman, and Pieter Abbeel.
\newblock Benchmarking deep reinforcement learning for continuous control.
\newblock In {\em ICML}, pages 1329--1338, 2016. 
\newblock {\em arXiv:1604.06778}.

\bibitem[\protect\citeauthoryear{Glorot and
  Bengio}{2010}]{glorot2010understanding}
Xavier Glorot and Yoshua Bengio.
\newblock Understanding the difficulty of training deep feedforward neural
  networks.
\newblock In {\em AISTATS}, pages 249--256, 2010.

\bibitem[\protect\citeauthoryear{Kingma and Ba}{2015}]{kingma2014adam}
Diederik~P Kingma and Jimmy Ba.
\newblock Adam: A method for stochastic optimization.
\newblock In {\em ICLR}, 2015.
\newblock {\em arXiv:1412.6980}.

\bibitem[\protect\citeauthoryear{Mania \bgroup \em et al.\egroup
  }{2018}]{mania2018simple}
Horia Mania, Aurelia Guy, and Benjamin Recht.
\newblock Simple random search provides a competitive approach to reinforcement
  learning.
\newblock {\em arXiv:1803.07055}, 2018.

\bibitem[\protect\citeauthoryear{Mnih \bgroup \em et al.\egroup
  }{2015}]{Mnih2015}
Volodymyr Mnih, Koray Kavukcuoglu, David Silver, Andrei~A. Rusu, Joel Veness,
  Marc~G. Bellemare, Alex Graves, Martin Riedmiller, Andreas~K. Fidjeland,
  Georg Ostrovski, Stig Petersen, Charles Beattie, Amir Sadik, Ioannis
  Antonoglou, Helen King, Dharshan Kumaran, Daan Wierstra, Shane Legg, and
  Demis Hassabis.
\newblock {Human-level control through deep reinforcement learning}.
\newblock {\em Nature}, 518(7540):529--533, 2015.

\bibitem[\protect\citeauthoryear{Osband \bgroup \em et al.\egroup
  }{2016}]{osband2016deep}
Ian Osband, Charles Blundell, Alexander Pritzel, and Benjamin Van~Roy.
\newblock Deep exploration via bootstrapped dqn.
\newblock In {\em NIPS}, pages 4026--4034, 2016.
\newblock {\em arXiv:1602.04621}.

\bibitem[\protect\citeauthoryear{Sutton and
  Barto}{1998}]{sutton1998reinforcement}
Richard~S Sutton and Andrew~G Barto.
\newblock {\em Reinforcement learning: An introduction}.
\newblock MIT press Cambridge, 1998.

\bibitem[\protect\citeauthoryear{Van~Hasselt \bgroup \em et al.\egroup
  }{2016}]{vanhasselt2016deep}
Hado Van~Hasselt, Arthur Guez, and David Silver.
\newblock Deep reinforcement learning with double q-learning.
\newblock In {\em AAAI}, volume~16, pages 2094--2100, 2016.
\newblock {\em arXiv:1509.06461}.

\end{thebibliography}

\end{document}